# Leveraging Large Language Models for Knowledge-free Weak Supervision in Clinical Natural Language Processing

Enshuo Hsu[1, 2, 3], MS; Kirk Roberts[1], PhD

[1]McWilliams School of Biomedical Informatics, University of Texas Health Science Center at Houston, Houston, Texas, USA.

[2]Center for Health Data Science & Analytics, Houston Methodist, Houston, TX, USA.

[3]Enterprise Dev & Integration, University of Texas MD Anderson Cancer Center, Houston, TX, USA.

Corresponding author: Kirk Roberts, kirk.roberts@uth.tmc.edu

**Word Count:**

Total: 4289

Abstract: 177

# ABSTRACT

The performance of deep learning-based natural language processing systems is based on large amounts of labeled training data which, in the clinical domain, are not easily available or affordable. Weak supervision and in-context learning offer partial solutions to this issue, particularly using large language models (LLMs), but their performance still trails traditional supervised methods with moderate amounts of gold-standard data. In particular, inferencing with LLMs is computationally heavy. We propose an approach leveraging fine-tuning LLMs and weak supervision with virtually no domain knowledge that still achieves consistently dominant performance. Using a prompt-based approach, the LLM is used to generate weakly-labeled data for training a downstream BERT model. The weakly supervised model is then further fine-tuned on small amounts of gold standard data. We evaluate this approach using Llama2 on three different n2c2 datasets. With no more than 10 gold standard notes, our final BERT models weakly supervised by fine-tuned Llama2-13B consistently outperformed out-of-the-box PubMedBERT by 4.7% to 47.9% in F1 scores. With only 50 gold standard notes, our models achieved close performance to fully fine-tuned systems.

# Introduction

Deep learning-based natural language processing (NLP) has achieved remarkable success in the open domain. However, achieving optimal performance in the clinical domain faces many challenges. First, training such complex architectures often requires a large labeled corpus [1]. Second, specific subpopulations (e.g., rare diseases, minority ethnicities) are often under-represented in clinical notes [2], magnifying the consequences of underpowered datasets. Third, even with sufficient notes available in electronic health records (EHRs), the protection of patient privacy makes access to the corpus challenging. Finally, manual annotation of a gold standard is not only a labor-intensive task, it also requires advanced clinical knowledge for interpretation of the text in clinical notes [3,4]. In recent years, approaches including weak supervision and in-context learning have been developed to address this challenge [1,4].

Weak supervision, which utilizes labeling functions (LFs) to generate noisy weak labels for model training, has already been adopted in the clinical domain [3,5–11]. Despite its promise, weak supervision still requires significant resources to construct LFs. The rule-based approach requires domain experts to handcraft decision rules [5–9]. The ontology-based approach requires that the concepts of interest be included in existing ontologies or dictionaries [10,11]. Data programming requires significant efforts from programmers who have a thorough understanding of the clinical data [3].

In-context learning, in which pre-trained large language models (LLMs) are prompted to predict textual outputs, is a relatively new method. In theory, it requires few ("few-shot") or even no ("zero-shot") training data [12,13]. However, recent studies raised concerns about underperformance [14–16] and instability [17] in the medical domain. Despite the appealing idea, at this point, there is no strong evidence to support the use of in-context learning as the frontline approach in a medical NLP system. Furthermore, due to the model sizes (measured as the number of parameters), LLM inference requires significant computation resources. We estimate that performing in-context learning with Llama2-13B, a 13 billion-parameter model [18] for 2018 i2b2 benchmark [19] (a subset of 505 discharge summaries from the MIMIC-III dataset [20]) requires $3.3 \times 10^{12}$ float point operations (FLOPs) per input sentence. On the other hand, predicting with a Bidirectional Encoder Representations from Transformers (BERT) model with 110 million parameters only requires $4.4 \times 10^{10}$ FLOPs per input sentence. This computational difference results in a dramatic difference in GPU time such that inferencing the entire collection of MIMIC-III discharge summaries would take an estimated 727 days on an NVIDIA A100 GPU while predicting with BERT would only take around 18 hours (Figure 2).

Recently, a few attempts have been made to combine the benefits of both weak supervision and in-context learning [21,22]. However, to our knowledge, there is no evaluation of an end-to-end approach in the medical domain that prompts an LLM for weak supervision and fine-tunes smaller models on the downstream task gold standard. The benefits and limitations of this method in a practical scenario where a small number of annotated notes are available have not been evaluated. Furthermore, fine-tuning LLM which has shown significant benefits in recent studies [23] has not been considered in such pipelines. Therefore, we propose an LLM-powered weak supervision approach that 1) minimizes domain expertise for rule-crafting and data programming and removes the dependency for ontologies by using the LLM to create weak labels, 2) leverages the latest prompt-based supervised fine-tuning (SFT) techniques to fine-tune LLMs, 3) consistently achieves dominant performances by weakly supervising and fine-tuning BERT [24] models for downstream tasks, and 4) avoids the computational burden of deploying LLMs in the production environment.

In this study, we evaluated four experimental settings as detailed in Table 1. The primary method, **Llama-SFT$n$-WS-BERT$n$** starts with supervised fine-tuning (SFT) Llama2-13B with a certain number ($n$) of gold standard notes in the training set. The fine-tuned Llama model then performs in-context learning on the rest of the training set to generate weak labels. We use the weak labels to perform weak supervision (WS) on BERT, followed by final fine-tuning with gold standards. Considering the high GPU memory requirement of SFT, we also proposed a compact version, **Llama-WS-BERT$n$** which the SFT of Llama2 was omitted. We use Llama2 out-of-the-box to perform weak supervision on BERT. For comparison, we evaluated two baselines, **Llama-SFT$n$** and **BERT$n$** which Llama2-13B and PubMedBERT were fine-tuned with $n$ gold standard notes. Details are described in the Methods section.

We evaluated three widely used biomedical benchmarks, the 2012 [25], 2014 [26], and 2018 [19] Integrating Biology and the Bedside (i2b2) Natural Language Processing Challenges for temporal relation extraction, protected health information (PHI) de-identification, and adverse drug events (ADEs) and medication properties extraction, respectively.

This study demonstrates a robust usage of LLMs that requires minimal to zero human input while achieving significant improvement in well-established benchmarks. We hypothesize that this approach is a safe and effective means of augmenting existing supervised clinical NLP approaches by inserting this simple technique between the now-standard pre-training and fine-tuning steps.

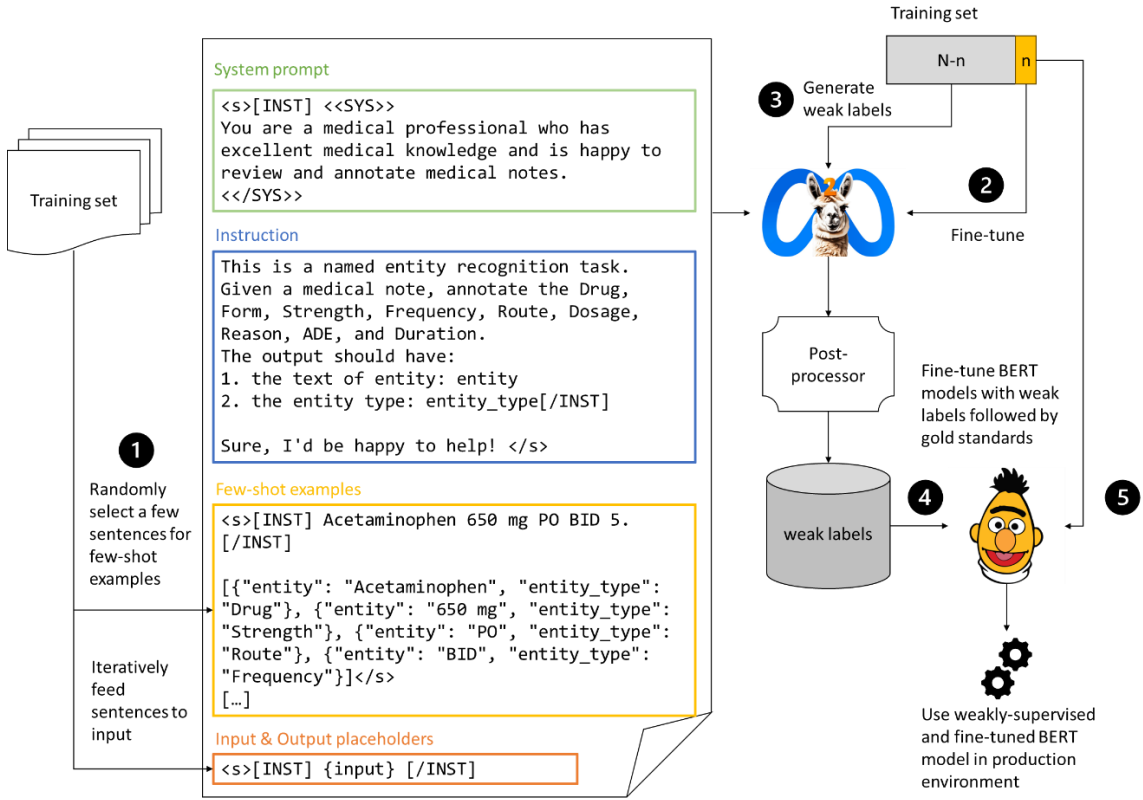

*Figure 1: Methodology flowchart. A prompt template is constructed with a few random sentences from the training set as the few-shot examples. For certain annotated gold standard notes, we first fine-tune Llama2-13B, then use the fine-tuned model to perform in-context learning to weakly supervise a PubMedBERT. Finally, we fine-tune the BERT model with gold standard notes and use it in the production environment.*

*Table 1: Experimental settings*

|  | Notation | Description | Product |
| --- | --- | --- | --- |
| Proposed methods | Llama-SFT*n*-WS-BERT*n* | Llama2-13B is supervised fine-tuned (SFT) with *n* gold standard notes in the training set, then performs few-shot in-context learning to generate weak labels. Weakly supervise BERT and fine-tune BERT with *n* gold standard notes. | A weakly supervised and fine-tuned BERT |
| **Baselines** | Llama-WS-BERT*n* | Llama2-13B out-of-the-box performs few-shot in-context learning to generate weak labels. Weakly supervise BERT and fine-tune BERT with *n* gold standard notes. | A weakly supervised and fine-tuned BERT |
| | Llama-SFT*n* | Llama2-13B is supervised fine-tuned with *n* gold standard notes in training set. | A fine-tuned Llama |
| | BERT*n* | Fine-tune BERT with *n* gold standard notes. | A fine-tuned BERT |

## Results

**LLMs inferencing is computationally expensive**

On the 2018 benchmark, which contains a subset of 505 MIMIC-III discharge summaries, Llama2-13B spent 147 GPU hours in total, with a median of 16 minutes (Q1-Q3 = 11-22 minutes) to create weak labels on each note. Since the computation for inference is linear to the number of input instances, we fit a linear regression model to project the total GPU time for labeling all the 59,652 discharge summaries in the MIMIC-III dataset. The projected time on a single NVIDIA A100 GPU is 727 days. PubMedBERT took 9 minutes in total, with a median of 1 second (Q1-Q3 = 0.7-1.4 seconds) per note. The projected time for labeling all the discharge summaries in MIMIC-III is 18 hours and 16 minutes.

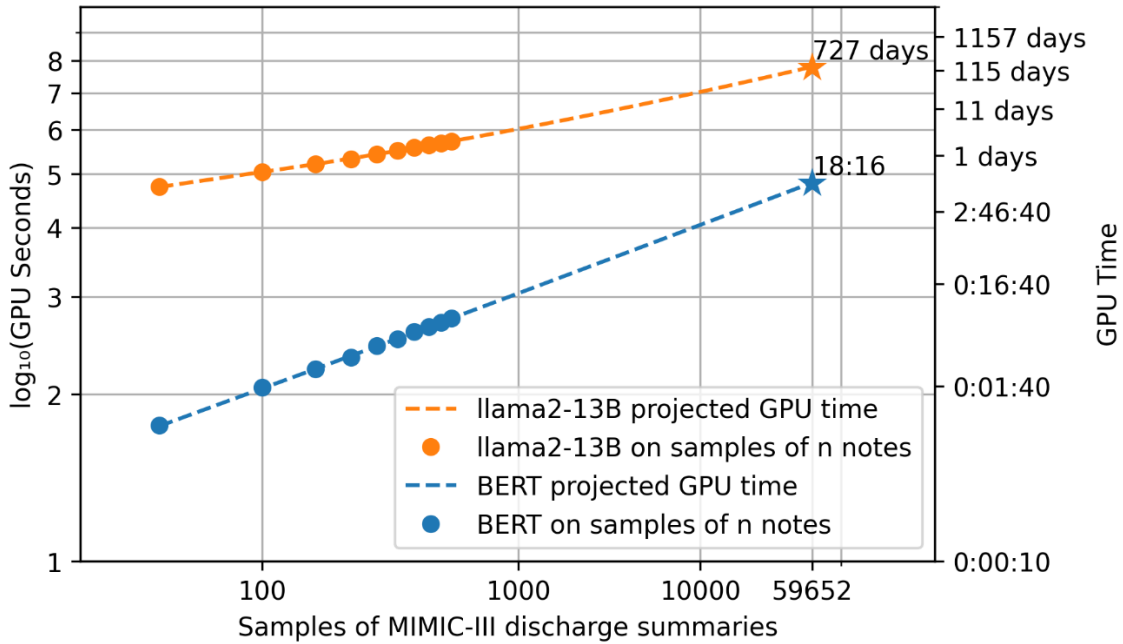

*Figure 2: Benchmarking GPU hours with MIMIC-III discharge summaries. The 505 discharge summaries in the 2018 i2b2 challenge were used to project the entire collection of discharge summaries. Running on an NVIDIA A100 GPU, Llama2-13B requires 727 days of GPU time, while PubMedBERT only requires about 18 hours.*

**LLM-generated weak labels**

For the 2012 benchmark, the out-of-the-box Llama2-13B and the fine-tuned Llama-SFT*3* generated weak labels with 9,804 and 20,402 entities, respectively. The median numbers of entities per sentence were 2 and 3, respectively. For the 2014 benchmark, Llama2-13B and Llama-SFT*3* generated weak labels with 18,062 and 15,190 entities, respectively, with a median of 1 entity per sentence. For the 2018 benchmark, Llama2-13B and Llama-SFT3 generated weak labels with 53,177 and 56,169 entities with 1 and 4 entities per sentence, respectively. Our post-processing algorithm was able to handle the majority of LLM predictions, with less than 1% of sentences failing due to inconsistent output formats (Table 2).

*Table 2: Summary of LLM-generated weak labels*

| Benchmarks (Training set) | 2012 | 2014 | 2018 |
|---|---|---|---|
| **Features** | | | |
| Notes | 190 | 790 | 303 |
| Sentences | 5,995 | 34,101 | 46,228 |
| Total entities | 17,933 | 17,401 | 50,951 |
| Entities per sentence, median [Q1, Q3] | 3 [2, 4] | 0 [0, 0] | 0 [0, 0] |
| Entities per sentence, mean (Std Dev) | 3.14 (2.34) | 0.51 (1.72) | 1.1 (3.13) |
| Entities per note, median [Q1, Q3] | 82 [56, 120] | 18 [13, 27] | 147 [96, 224] |
| **Weak labels generated by Llama2-13B** | | | |
| Post-processing failed, sentences (%) | 2 | 2 | 22 |
| Total entities | 9,804 | 18,062 | 53,177 |
| Entities per sentence, median [Q1, Q3] | 2 [1, 3] | 1 [1, 1] | 1 [1, 3] |
| Entities per note, median [Q1, Q3] | 46 [31, 61] | 19 [12, 29] | 165 [102, 232] |
| **Weak labels generated by Llama-SFT$3$** | | | |
| Post-processing failed, sentences (%) | 14 | 27 | 82 |
| Total entities | 20,402 | 15,190 | 56,169 |
| Entities per sentence, median [Q1, Q3] | 3 [2, 5] | 1 [1, 4] | 4 [2, 7] |
| Entities per note, median [Q1, Q3] | 88 [59, 132] | 15 [10, 21] | 181 [111, 246] |
| **Weak labels generated by Llama-SFT$5$** | | | |
| Post-processing failed, sentences (%) | 76 | 34 | 48 |
| Total entities | 18,676 | 15,107 | 48,143 |
| Entities per sentence, median [Q1, Q3] | 3 [2, 5] | 1 [1, 3] | 4 [2, 7] |
| Entities per note, median [Q1, Q3] | 82 [57, 119] | 14 [9, 22] | 147 [90, 215] |
| **Weak labels generated by Llama-SFT$10$** | | | |
| Post-processing failed, sentences (%) | 69 | 28 | 25 |
| Total entities | 18,386 | 13,743 | 46,223 |
| Entities per sentence, median [Q1, Q3] | 3 [2, 5] | 1 [1, 2] | 4 [2, 7] |
| Entities per note, median [Q1, Q3] | 81 [54, 117] | 15 [11, 20] | 147 [84, 207] |
| **Weak labels generated by Llama-SFT$50$** | | | |
| Post-processing failed, sentences (%) | 72 | 54 | 59 |
| Total entities | 18,420 | 18,415 | 47,688 |
| Entities per sentence, median [Q1, Q3] | 3 [2, 5] | 1 [1, 3] | 4 [2, 7] |
| Entities per note, median [Q1, Q3] | 77 [57, 118] | 17 [12, 25] | 150 [87, 212] |

**Proposed method: Llama-SFT$n$-WS-BERT$n$**

Our primary proposed method, Llama-SFT$n$-WS-BERT$n$ consistently achieved dominant performance in most experiments across the three benchmarks. In the extremely low-resourced setting in which only 3 gold standard notes were used, on the 2012 events benchmark, time expression benchmark, 2014 benchmark, and 2018 benchmark, Llama-SFT*3*-WS-BERT*3* achieved F1 scores of 0.7765, 0.7538, 0.6336, and 0.7747, while the baseline Llama-SFT*3* had 0.7418, 0.6045, 0.5898, and 0.6252; BERT3 had 0.5953, 0.2753, 0.3083, and 0.6555. Llama-SFT*3*-WS-BERT*3* outperformed the Llama-SFT*3* baseline by 3.5% to 15.0% and the BERT*3* baseline by 11.9% to 47.9% in the F1 score. When 10 gold standard notes were used, Llama-SFT*10*-WS-BERT*10* achieved F1 scores of 0.8466, 0.8448, 0.6942, and 0.8005, which is 3.2% to 14.6% higher than the Llama-SFT*10* baseline and 4.7% to 16.8% higher than the BERT*10* baseline. In the relatively annotation-abundant scenario when 50 gold standard notes were used, the Llama-SFT*50*-WS-BERT*50* achieved close performance to fully supervised BERT models by only 2.8%, 2.5%, 6.1%, and 2.2% lower in F1 score. For the 2012 time expression benchmark, however, the F1 score of Llama-SFT*50*-WS-BERT*50* is slightly lower than BERT*50* by 1.3%.

**Proposed method: Llama-WS-BERT*n***

The compact method Llama-WS-BERT*n* showed improved performance in most benchmarks. On the 2012 event benchmark, Llama-WS-BERT*n* and Llama-SFT*n* had similar performance and the differences are 0.5% to 2.5% for *n* from 3 to 50. While it outperformed the BERT*n* by up to 17.1%. In the 2012 temporal expression benchmark, Llama-WS-BERT*n* and Llama-SFT*n* had similar performances when *n* was less than 10. While Llama-WS-BERT*10* and Llama-WS-BERT*50* outperformed Llama-SFT*10* and Llama-SFT*50* by 7.9% and 13.5%, respectively. On the 2014 benchmark, Llama-WS-BERT*n* and Llama-SFT*n* had similar performances except for *n* of 5. Llama-WS-BERT*n* outperformed BERT*n* by 3.5% to 27.5%. On the 2018 benchmark, Llama-WS-BERT*n* outperformed Llama-SFT*n* and BERT*n* by 5.6% to 11.8% and 1.1% to 8.8%, respectively. Overall, Llama-WS-BERT*n* performs similar to or better than the Llama-SFT*n* baseline while dominating the BERT*n* baseline on most benchmarks.

**Baseline methods**

On the 2012 benchmarks, under the low-resource setting ($n < 10$), Llama-SFT$n$ performed better than BERT$n$. While when $n = 50$, BERT$50$ outperformed Llama-SFT$50$. On the 2014 benchmark, Llama-SFT$n$ outperformed BERT$n$ across the board by 5% to 28.2%. On the 2018 benchmark, BERT$n$ outperformed Llama-SFT$n$ by 1.5% to 7.4%.

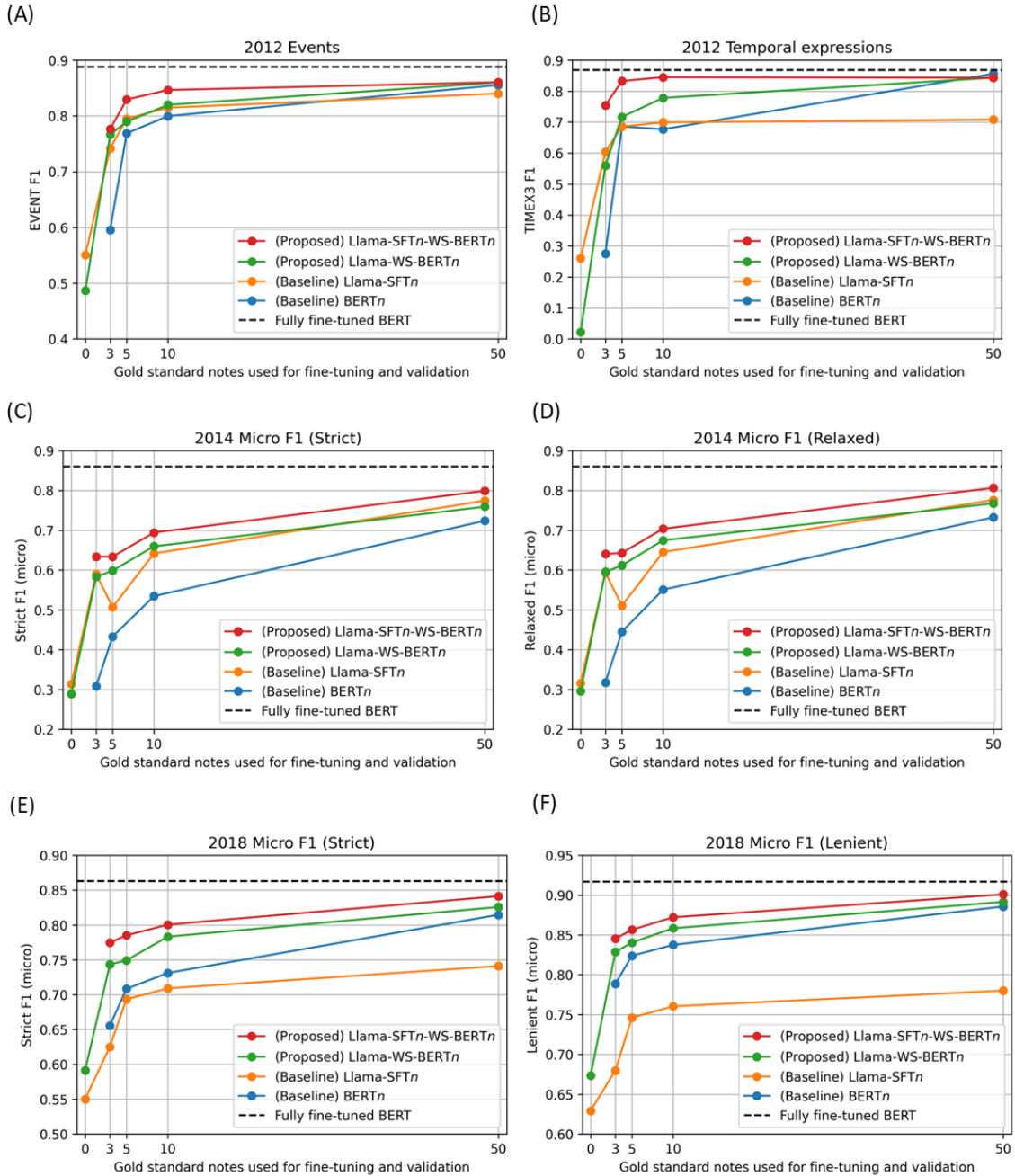

Figure 3: Weakly supervised end models fine-tuned on 3, 5, 10, and 50 gold standard notes from the training set compared to BERT models without weak supervision. (A) 2012 i2b2 challenge events extraction F1 score and (B) temporal expression extraction F1 score. (C) 2014 i2b2 challenge Strict micro F1 score and (D) Relaxed micro F1 score. (E) 2018 i2b2 challenge Strict micro F1 score and (F) Lenient micro F1 score.

## Discussion

We proposed an LLM-powered weak supervision system that costs minimal to zero domain knowledge to improve the performance of clinical information extraction by 4.7% to 47.9% from the BERT baseline when no more than 10 gold standard notes were used for training. When 50 gold standard notes were used, our system achieved similar performance as a fully supervised BERT with a 2.2% to 6.1% difference. The method showed an overall benefit of fine-tuning low training sizes across the three benchmarks. Considering the computational burden of fine-tuning LLMs, we also proposed a compact version using Llama2 out-of-the-box and achieved improved performances across the board. The products of our methods are fine-tuned BERT models with 110 million parameters. Compared to modern LLMs which often have billions of parameters, the compact size makes model deployment more computationally efficient. Our framework (i.e., LLM, SFT, prompt templates, and post-processing algorithms) is domain-independent and can be applied to most medical information extraction systems. We expect the performance of this framework will improve further when more medically-focused LLMs become available. We conclude that the proposed method is a generalizable and effortless booster for low-training-size scenarios.

This study is one of the early works exploring the potential use of LLMs in the medical domain. Recent studies have debated the feasibility and performance of in-context learning for information extraction [14,15,17,27]. Following the ideas of LLM-powered labeling functions [28] and clinical knowledge distillation [21], we proposed a robust alternative that combines supervised fine-tuning LLMs, in-context learning, and weak supervision to achieve stably dominant performances. As a knowledge-free alternative for labeling functions, our study also points out a direction in which current weak supervision methods could be free from the heavy reliance on domain expert inputs and ontology.

On the 2012 time expression benchmark, when 50 the gold standard notes were used for training, our Llama-SFT$_{50}$-WS-BERT$_{50}$ had slightly reduced performances by 1.3% compared to the BERT$_{50}$ baseline. This finding is consistent with a recent weak supervision study which showed negative impact when a large amount of training notes were provided [3]. The most likely explanation is that when gold standards are adequate to provide the model with correct knowledge, the noise in the weak labels exceed the benefits. However, the performance drops in such cases with our approach are quite small, suggesting such an approach can have endurance upsides with little chances of catastrophic loss, unlike other LLM use cases.

On the 2014 and the 2018 benchmarks, we observed reversed results between the two baselines Llama-SFT$_n$ and BERT$_n$ in which Llama-SFT$_n$ performed better on the 2014 PHI de-identification task while BERT$_n$ performed better on the 2018 ADE & medication extraction task. One explanation is that since Llama2 is a general-domain model while PubMedBERT is a biomedical model, the former might have advantages in solving non-medical problems such as PHI identification while the latter has advantages in solving medical problems like medication terms.

Despite the promising results, this study does have a few limitations. First, unlike other weak supervision studies in which a large number of unlabeled notes were processed by LFs [2,3], for computational considerations, we chose benchmarks with relatively small sample sizes. We would expect that with larger weakly-labeled datasets the performance of our approach should increase, though this requires further experimentation. However, even with less than 800 notes, the LLM was able to generate weak labels that dramatically improved performance. Second, as an initial work, we did not evaluate different LLMs. We selected Llama2-13B based on the reported performances in the medical domain and their open-source and lightweight features [29]. Other open-source LLMs should be evaluated in future studies. Third, to keep the study focused, we did not evaluate different settings in supervised fine-tuning (e.g., prompt templates, learning rate), in-context learning (e.g., prompt templates, the number of few-shot examples), post-processing (e.g., label harmonization), and BERT model fine-tuning. We follow reported best practices for those [12,14,18]. We expect the performance to further improve if those details are carefully tuned.

## Conclusion

In conclusion, we proposed a novel method that combines LLMs and weak supervision for high-performance medical information extraction while minimizing domain knowledge dependence. Our method shows a consistent benefit. Further performance improvements are anticipated with more refined in-context learning and fine-tuning.

## Methods

Figure 1 provides an overview of our approach. We first constructed a **prompt template** with a system prompt, an instruction, few-shot examples sampled from the training set, and an input/output placeholder. For a given set of *n* gold standard notes, we fine-tuned Llama2-13B via prompt-based **supervised fine-tuning (SFT)**. We then used the fine-tuned Llama2 for **few-shot in-context learning** on the unannotated notes to generate weak labels. The weak labels were used to fine-tune ("**weakly supervise**") a BERT model. The BERT model was then **fine-tuned** with the gold standard notes to achieve optimal performance.

**Benchmarks**

We used datasets and tasks from 2012 [25], 2014 [26], and 2018 [19] Integrating Biology and the Bedside (i2b2) Natural Language Processing Challenge as benchmarks.

The 2012 i2b2 challenge focused on temporal relation extraction with 310 annotated clinical notes. Entities include 1) clinically significant events ("EVENT"), such as problems, tests, treatments, clinical departments, admissions, and transfers between departments, and 2) temporal expressions ("TIMEX3"), which are dates, times, durations, or frequencies phrases. For this study, the F1 scores for events and time expressions are used as the main metrics, while the temporal relations between events and time expressions are not evaluated.

The 2014 i2b2 challenge de-identification track focused on extracting Health Insurance Portability and Accountability Act (HIPAA) protected health information (PHI) from 1304 annotated clinical notes. We used the i2b2-PHI entities which include 7 types of PHI. We used strict and relaxed micro F1 scores as the main metrics.

The 2018 i2b2 challenge track 2 focused on the extraction of adverse drug events (ADEs) and medication properties from 505 discharge notes. The concept extraction task defined 9 entity types: drug, strength, form, dosage, frequency, route, duration, reason, and ADE. We used the Strict and Lenient micro F1 as the main metrics.

**Prompt templates**: We prepared a prompt template for each benchmark task as highlighted in Figure 1 and listed in Table S1. Our design adopted recent studies in prompt engineering [14,21] which includes 4 sections: 1) **system prompt**, in which a role is assigned to Llama2 to provide the context and to avoid triggering the safety features of the LLM, 2) **instruction**, which is a narrative description of the background (e.g., medical notes), task (e.g., named entity recognition, entity types), and expected output (i.e., the entity text and the entity type), 3) **few-shot examples**, where 8 randomly sampled sentences and the corresponding gold standard labels were listed following the JavaScript Object Notation (JSON) format. 4) **Input placeholders**, where for each sentence the text was placed in the input placeholder while the prompt was fed to LLM. The LLM would output text following the "[/INST]" special token which we collect for post-processing.

**Supervised fine-tuning (SFT) Llama2**

We used the prompt template described in the previous section to perform SFT. Each sentence in the gold standard notes was placed in the input placeholder and fed to the Llama2. Note that SFT is auto-regressive thus the labels were appended following the "[/INST]" special token after the input sentences. Following the original SFT hyperparameters[18], we use a cosine learning rate schedule with a $2 \times 10^{-5}$ initial learning rate and a weight decay of 0.1. The sequence length was 4096. We trained for 2 epochs. Due to the limiting GPU memory, we set the batch size to 1.

**Few-shot in-context learning**

LLMs have limitations in the number of input tokens due to their transformer architecture. Llama2-13B has a limit of 4096 input tokens. Including entire clinical notes in a prompt would often exceed the maximum input length. Therefore, we performed in-context learning at the sentence level. We sentence-segmented each note with spaCy 3.5.4 Sentencizer [30]. Sentences were placed in the input placeholder and the output was collected after the "[/INST]" special token for post-processing. To maximize the reproducibility, we set the top-k parameter to 1 which disabled random sampling of generated tokens. To increase the text generation speed, we set the maximum output length to 128 tokens.

**Large language models**: Clinical notes often include PHI and are restricted from sharing. LLMs that are only available through API (e.g., GPT-3, GPT-4) [17,31] could be limiting in real-world scenarios. An ideal LLM for our system meets the three criteria: 1) is open-source and can be deployed locally, 2) is lightweight enough for making inferences on a local server, and 3) has high performance in the medical domain. Llama2 is a pre-trained open-source large language model that comes with different sizes of architecture from 7 billion to 65 billion parameters and has demonstrated competitive performances in both open-domain and biomedical NLP benchmarks [29]. The 7 billion parameter version ("Llama2-7B") loaded in 16-bit floating-point can fit in a GPU with 14 GB of vRAM, while the 13 billion parameter version ("Llama2-13B") fits in 26 GB of vRAM. We chose Llama2-13B for a balance of performance and computation cost.

**Post-processing**: To serve the purpose of minimizing human effort, our post-processing was designed to be automatic, robust, and generalizable across tasks. The steps were: 1) **generated-text extraction**, which extracts all generated text after the "[/INST]" special token. In cases where excessive text was generated after the intended JSON format, for instance, a new "[INST]" was generated by Llama2, we truncated it. 2) **JSON formatting,** which is a simple regular expression logic that extracts the "\{.*?\}" patterns in a JSON list. 3) **Entity recovery**, which utilized the extracted entity text to identify the span in the input sentence. 4) **Entity type filtering,** which filters out irrelevant entity types that Llama2 created and are not one of the entity types for the benchmark tasks. We used exact, case-sensitive string matching to minimize potential bias from human interpretation. By the end of post-processing, we obtained a list of entities with the span, entity text, and entity type for each clinical note.

**Weak supervision**

We used one of the latest state-of-the-art biomedical BERT models, PubMedBERT [32] (denoted as BERT) in this study. To evaluate the scenario where only a few annotated notes are available for training, the BERT model was first fine-tuned with weak labels from *(N-$n_s$)* notes followed by fine-tuning with gold labels from $n_s$ notes, where *N* is the total number of training notes, $n_s \in \{3, 5, 10, 50\}$. To ensure the $n_s$ notes were representative, they were selected such as having the closest number of entities to the median number of entities among all notes in the official training set. The formula below defines the selected subset $S_{n_s}$:

$$S_{n_s} = \{note_i : i \text{ in top } n_s \, argmin(abs(\# \, of \, entities \, in \, note_i - median \, \# \, of \, entities))\}$$

**Fine-tuning BERT**

Fine-tuning with weak labels and gold standard data follows similar methods, with a few differences in hyperparameters (Table S2). To segment notes into shorter chunks that the BERT models could process, we sentence-segmented the notes with spaCy. For each sentence, word tokenization was performed using the WordPiece algorithm implemented in the Python *transformers* module (version 4.30.2) and based on a pre-defined dictionary.

For fine-tuning, the development set was divided into a training set (80%) and a validation set (20%), unless specified in Table S2. Model weights were saved as checkpoints after each training period ("epoch"), and optimal checkpoint weights were selected during validation as our final NLP model. For efficiency, an early stop criterion of 8 continuous non-improving epochs was used. The NLP models were implemented using Python 3.9.7, *PyTorch* 2.0.1, and *transformers* 4.30.2. All computations were performed on a server with 8 NVIDIA A100 80GB GPU.

**Benchmarking FLOPs and GPU time**

The corpus in the 2018 benchmark is a subset of 505 discharge summaries from the MIMIC-III [20] database. We calculated the FLOPs for inferencing one input sentence with Llama2-13B following the formula below [33],

$$N_{tokens}(2N + 2n_{layer}n_{ctx}d_{attn})$$

$$= 128 \times (2 \times (13015864320) + 2 \times 40 \times 400 \times 4096)$$

$$\cong 3.348 \times 10^{12}$$

where $N_{tokens}$ denotes the number of tokens Llama2 outputs; $N$ denotes the total parameters in the model; $n_{layer}$ denotes the number of layers in the model; $n_{ctx}$ denotes the input context token length. We use the length of the prompt template to estimate. $d_{attn}$ denotes the dimension of attention output. We monitored the FLOPs for PubMedBERT with the built-in tool, *profiler* in *PyTorch*. The GPU time for each note was monitored during the inferencing with Llama2-13B and the prediction with PubMedBERT. We randomly sampled 50 to 500 notes and fitted a linear regression line to model the correlation between the number of notes and the total GPU time. A projection was made to estimate the total GPU time required for all the discharge summaries from the MIMIC-III database.

## COMPETING INTERESTS

None

## FUNDING

This work was partially supported by awards from the National Institutes of Health, including the National Institute of Biomedical Imaging and Bioengineering (NIBIB: R21EB029575) and the National Institute of Allergy & Infectious Diseases (NIAID: R21AI164100).

**APPENDIX**

*Table S1: Prompt templates*

| | **Prompt template** |
|---|---|

```
<s>[INST] <<SYS>>
You are a medical professional who has excellent medical
knowledge and is happy to review and annotate medical notes.
<</SYS>>
This is a named entity recognition task. Given a medical note,
annotate the events (EVENT) and time expressions (TIMEX3):
The output should have:
1. the text of entity: entity
2. the entity type: entity_type[/INST]

Sure, I'd be happy to help! </s>

<s>[INST]  At 9/7/93 , 1:00 a.m. , intravenous fluids rate was
decreased to 50 cc's per hour , total fluids given during
the first 24 hours were 140 to 150 cc's per kilo per day .
[/INST]

[{"entity": "intravenous fluids", "entity_type": "EVENT"},
{"entity": "decreased", "entity_type": "EVENT"}, {"entity":
"total fluids", "entity_type": "EVENT"}, {"entity": "9/7/93 ,
1:00 a.m.", "entity_type": "TIMEX3"}, {"entity": "24 hours",
"entity_type": "TIMEX3"}] </s>

<s>[INST] At that time it appeared well adhered to the underlying
skin . [/INST]

[{"entity": "it", "entity_type": "EVENT"}, {"entity": "well
adhered", "entity_type": "EVENT"}] </s>

<s>[INST] LAD and LCX were stented , but PLB could not be crossed
by guidewire ( were collaterals ) .[/INST]

[{"entity": "stented", "entity_type": "EVENT"}, {"entity":
"guidewire", "entity_type": "EVENT"}] </s>

<s>[INST] Postoperatively , she diuresed very well .[/INST]

[{"entity": "diuresed", "entity_type": "EVENT"}] </s>

<s>[INST] Brief Hospital Course : Mr. Williams was admitted to
Hallmark Health System on 05-08 and taken to the operating room
with Dr. Barber for a CABGx2 and AVR w/ 25 mm pericardial
valve .[/INST]

[{"entity": "admitted", "entity_type": "EVENT"}, {"entity":
"Hallmark Health System", "entity_type": "EVENT"}, {"entity":
"the operating room", "entity_type": "EVENT"}, {"entity": "a
CABGx2", "entity_type": "EVENT"}, {"entity": "AVR",
"entity_type": "EVENT"}, {"entity": "25 mm pericardial valve",
```

"entity_type": "EVENT"}, {"entity": "05-08", "entity_type": "TIMEX3"}] </s>

<s>[INST] She continued to require pressor support in the high 30 's .[/INST]

[{"entity": "pressor support", "entity_type": "EVENT"}] </s>

<s>[INST] She was seen at A Memorial Hospital where her blood pressure was measured at 240/120 and was found to have a mild left hemiparesis .[/INST]

[{"entity": "A Memorial Hospital", "entity_type": "EVENT"}, {"entity": "her blood pressure", "entity_type": "EVENT"}, {"entity": "a mild left hemiparesis", "entity_type": "EVENT"}] </s>

<s>[INST] She had worsening abdominal pain on exam .[/INST]

[{"entity": "worsening abdominal pain", "entity_type": "EVENT"}, {"entity": "exam", "entity_type": "EVENT"}] </s>

<s>[INST] {input} [/INST]

```
<s>[INST] <<SYS>>
You are a medical professional who has excellent medical
knowledge and is happy to review and annotate medical notes.
<</SYS>>
This is a named entity recognition task. Given a medical note,
annotate the Protected Health Information (PHI):

1. NAME
      a. NAME_PATIENT
      b. NAME_DOCTOR
      c. NAME_USERNAME
2. PROFESSION
3. LOCATION
      a. LOCATION_HOSPITAL
      b. LOCATION_ORGANIZATION
      c. LOCATION_STREET
      d. LOCATION_CITY
      e. LOCATION_STATE
      f. LOCATION_COUNTRY
      g. LOCATION_ZIP
      h. LOCATION_LOCATION-OTHER
4. AGE
5. DATE
6. CONTACT
      a. CONTACT_PHONE
      b. CONTACT_FAX
      c. CONTACT_EMAIL
      d. CONTACT_URL
7. ID
      a. ID_BIOID
      b. ID_DEVICE
      c. ID_HEALTHPLAN
      d. ID_IDNUM
      e. ID_MEDICALRECORD

The output should have:
1. the text of entity: entity
2. the entity type: entity_type[/INST]

Sure, I'd be happy to help! </s>

<s>[INST] Medical Decision Making ED Course   75M w/acute on
chronic low back and left buttock pain, now improved after pain
medication in the ED. [/INST]

[{"entity": "75", "entity_type": "AGE"}] </s>

<s>[INST] 2079, Resumed oral hypoglycemic agent 11/86. [/INST]
```

[{"entity": "2079", "entity_type": "DATE"}, {"entity": "11/86", "entity_type": "DATE"}] </s>

<s>[INST] Ms. Yerger's postoperative  course was notable for persistent pleural effusion and transient  episode of atrial fibrillation. [/INST]

[{"entity": "Yerger", "entity_type": "NAME_PATIENT"}] </s>

<s>[INST] 3/06/2120 3:50 p.m.  Time paged:3:38 p.m.  Tine Called Back: 3:42 p.m.  Time Reccs given: 4:30 p.m.   Consulted by:  ED for L. facial droop, dysarthria; English speaking, interviewed with translator    HPI: Ms. Lindsay is 76y.o. [/INST]

[{"entity": "3/06/2120", "entity_type": "DATE"}, {"entity": "English", "entity_type": "LOCATION_COUNTRY"}, {"entity": "Lindsay", "entity_type": "NAME_PATIENT"}, {"entity": "76", "entity_type": "AGE"}] </s>

<s>[INST] On 9/25/75 in the morning he had a fall when getting out of bed and struck his head, and brought to the ED. [/INST]

[{"entity": "9/25/75", "entity_type": "DATE"}] </s>

<s>[INST] She will continue 3x/week dialysis in her Educare-Pullman outpatient unit. [/INST]

[{"entity": "Educare-Pullman", "entity_type": "LOCATION_HOSPITAL"}] </s>

<s>[INST] [vmf47] ENT:Patient's airway is intact. [/INST]

[{"entity": "vmf47", "entity_type": "NAME_USERNAME"}] </s>

<s>[INST] ROS:  No chest pain, change in bowels, trouble with urination      SH:  Lives with wife  Decorator for income  Neither using ETOH      EXAM:  BP 142/60 P 80 wt 174 lbs  Appears well Eyes- no lid lag, full motion, full fields, Neck- small thyroid, Lungs-  clear, Cor0RR S1 S2 S4, Abd- + [/INST]

[{"entity": "Decorator", "entity_type": "PROFESSION"}] </s>

<s>[INST] {input} [/INST]

<s>[INST] <<SYS>>
You are a medical professional who has excellent medical knowledge and is happy to review and annotate medical notes.
<</SYS>>
This is a named entity recognition task. Given a medical note, annotate the Drug, Form, Strength, Frequency, Route, Dosage, Reason, ADE, and Duration.
The output should have:
1. the text of entity: entity
2. the entity type: entity_type[/INST]

Sure, I'd be happy to help! </s>

<s>[INST] Acetaminophen 650 mg PO BID 5. [/INST]

[{"entity": "Acetaminophen", "entity_type": "Drug"}, {"entity": "650 mg", "entity_type": "Strength"}, {"entity": "PO", "entity_type": "Route"}, {"entity": "BID", "entity_type": "Frequency"}] </s>

<s>[INST] Mesalamine DR 1200 mg PO BID 2. [/INST]

[{"entity": "Mesalamine DR", "entity_type": "Drug"}, {"entity": "1200 mg", "entity_type": "Strength"}, {"entity": "BID", "entity_type": "Frequency"}, {"entity": "PO", "entity_type": "Route"}] </s>

<s>[INST] However, her symptoms were relieved with nitro. [/INST]

[{"entity": "nitro", "entity_type": "Drug"}] </s>

<s>[INST] Levofloxacin 500 mg Tablet Sig: One (1) Tablet PO Q24H (every 24 hours) for 2 days. [/INST]

[{"entity": "Levofloxacin", "entity_type": "Drug"}, {"entity": "500 mg", "entity_type": "Strength"}, {"entity": "Tablet", "entity_type": "Form"}, {"entity": "One (1)", "entity_type": "Dosage"}, {"entity": "Tablet", "entity_type": "Form"}, {"entity": "PO", "entity_type": "Route"}, {"entity": "Q24H", "entity_type": "Frequency"}, {"entity": "for 2 days", "entity_type": "Duration"}] </s>

<s>[INST] [**Hospital1 **] COUMADIN 5 mg daily DIGOXIN .0625 mg daily DONEPEZIL 10 mg daily FERROUS SULFATE 325 mg daily FUROSEMIDE 60 mg daily INSULIN NPH - 12 units once a day INSULIN LISPRO [/INST]

[{"entity": "COUMADIN", "entity_type": "Drug"}, {"entity": "5 mg", "entity_type": "Strength"}, {"entity": "daily", "entity_type": "Frequency"}, {"entity": "DIGOXIN", "entity_type":

*2018 (Track 2) ADE and Medication Extraction Challenge*

"Drug"}, {"entity": ".0625 mg", "entity_type": "Strength"}, {"entity": "daily", "entity_type": "Frequency"}, {"entity": "DONEPEZIL", "entity_type": "Drug"}, {"entity": "10 mg", "entity_type": "Strength"}, {"entity": "daily", "entity_type": "Frequency"}, {"entity": "FERROUS SULFATE", "entity_type": "Drug"}, {"entity": "325 mg", "entity_type": "Strength"}, {"entity": "daily", "entity_type": "Frequency"}, {"entity": "FUROSEMIDE", "entity_type": "Drug"}, {"entity": "60 mg", "entity_type": "Strength"}, {"entity": "daily", "entity_type": "Frequency"}, {"entity": "INSULIN NPH", "entity_type": "Drug"}, {"entity": "12", "entity_type": "Dosage"}, {"entity": "units", "entity_type": "Form"}, {"entity": "once a day", "entity_type": "Frequency"}, {"entity": "INSULIN LISPRO", "entity_type": "Drug"}] </s>

<s>[INST] [**2179**] Glipizide 10 mg po bid start [**2179**] Actos 30mg po daily(changed by his PCP in [**Month (only) **] from 15mg) Verapamil sr 240mg po daily start [**2173**] Lisinopril 40 mg po daily Lovastin 10mg po daily start [**2182**] [/INST]

[{"entity": "Glipizide", "entity_type": "Drug"}, {"entity": "10 mg", "entity_type": "Strength"}, {"entity": "po", "entity_type": "Route"}, {"entity": "bid", "entity_type": "Frequency"}, {"entity": "Actos", "entity_type": "Drug"}, {"entity": "30mg", "entity_type": "Strength"}, {"entity": "po", "entity_type": "Route"}, {"entity": "daily", "entity_type": "Frequency"}, {"entity": "Verapamil sr", "entity_type": "Drug"}, {"entity": "240mg", "entity_type": "Strength"}, {"entity": "po", "entity_type": "Route"}, {"entity": "daily", "entity_type": "Frequency"}, {"entity": "Lisinopril", "entity_type": "Drug"}, {"entity": "40 mg", "entity_type": "Strength"}, {"entity": "po", "entity_type": "Route"}, {"entity": "daily", "entity_type": "Frequency"}, {"entity": "Lovastin", "entity_type": "Drug"}, {"entity": "10mg", "entity_type": "Strength"}, {"entity": "po", "entity_type": "Route"}, {"entity": "daily", "entity_type": "Frequency"}] </s>

<s>[INST] The patient apparently had an undocumented drug side effect of the lorazepam, which resulted in two code purples being called. [/INST]

[{"entity": "lorazepam", "entity_type": "Drug"}] </s>

<s>[INST] Lasix 20 mg Tablet Sig: Three (3) Tablet PO once a day: This medication should not be restarted until the patient is evaluated by a physician at the nursing home. [/INST]

[{"entity": "Lasix", "entity_type": "Drug"}, {"entity": "20 mg", "entity_type": "Strength"}, {"entity": "Tablet", "entity_type":

|  |  |
|---|---|
| "Form"}, {"entity": "Three (3)", "entity_type": "Dosage"}, {"entity": "Tablet", "entity_type": "Form"}, {"entity": "PO", "entity_type": "Route"}, {"entity": "once a day", "entity_type": "Frequency"}] </s><br><br><s>[INST] {input} [/INST] |

To be compliant with the data user agreement, the few-shot example sentences in this table were intentionally masked with [Example sentence #1] to [Example sentence #8]. The actual sentence text was used in the prompts.

*Table S2: Hyperparameters in fine-tuning*

| Run | Validation ratio | Input token length | Learning rate | Batch size |
|---|---|---|---|---|
| **2012 Temporal Relations Challenge** | | | | |
| Bert_gold10 | 0.2 | 256 | 2.00E-06 | 2 |
| Bert_gold100% | 0.2 | 256 | 2.00E-06 | 32 |
| Bert_gold3 | 0.34 | 256 | 2.00E-06 | 1 |
| Bert_gold5 | 0.2 | 256 | 2.00E-06 | 2 |
| Bert_gold50 | 0.2 | 256 | 2.00E-06 | 2 |
| llama2-13B_Bert_gold0 | 0.2 | 256 | 2.00E-06 | 32 |
| llama2-13B_Bert_gold10 | 0.2 | 256 | 2.00E-06 | 2 |
| llama2-13B_Bert_gold10_ws | 0.2 | 256 | 2.00E-06 | 32 |
| llama2-13B_Bert_gold3 | 0.34 | 256 | 2.00E-06 | 1 |
| llama2-13B_Bert_gold3_ws | 0.2 | 256 | 2.00E-06 | 32 |
| llama2-13B_Bert_gold5 | 0.2 | 256 | 2.00E-06 | 2 |
| llama2-13B_Bert_gold50 | 0.2 | 256 | 2.00E-06 | 2 |
| llama2-13B_Bert_gold50_ws | 0.2 | 256 | 2.00E-06 | 32 |
| llama2-13B_Bert_gold5_ws | 0.2 | 256 | 2.00E-06 | 32 |
| llama2-13B_gold10_Bert_gold10 | 0.2 | 256 | 2.00E-06 | 2 |
| llama2-13B_gold10_Bert_gold10_ws | 0.2 | 256 | 2.00E-06 | 32 |
| llama2-13B_gold3_Bert_gold3 | 0.34 | 256 | 2.00E-06 | 1 |
| llama2-13B_gold3_Bert_gold3_ws | 0.2 | 256 | 2.00E-06 | 32 |
| llama2-13B_gold50_Bert_gold50 | 0.2 | 256 | 2.00E-06 | 2 |
| llama2-13B_gold50_Bert_gold50_ws | 0.2 | 256 | 2.00E-06 | 32 |
| llama2-13B_gold5_Bert_gold5 | 0.2 | 256 | 2.00E-06 | 2 |
| llama2-13B_gold5_Bert_gold5_ws | 0.2 | 256 | 2.00E-06 | 32 |

| 2014 De-identification Challenge | | | | |
|---|---|---|---|---|
| Bert_gold10 | 0.2 | 256 | 2.00E-06 | 2 |
| Bert_gold100% | 0.2 | 256 | 2.00E-06 | 32 |
| Bert_gold3 | 0.34 | 256 | 2.00E-06 | 1 |
| Bert_gold5 | 0.2 | 256 | 2.00E-06 | 2 |
| Bert_gold50 | 0.2 | 256 | 2.00E-06 | 2 |
| llama2-13B_Bert_gold0 | 0.2 | 256 | 2.00E-06 | 32 |
| llama2-13B_Bert_gold10 | 0.2 | 256 | 2.00E-06 | 2 |
| llama2-13B_Bert_gold10_ws | 0.2 | 256 | 2.00E-06 | 32 |
| llama2-13B_Bert_gold3 | 0.34 | 256 | 2.00E-06 | 1 |
| llama2-13B_Bert_gold3_ws | 0.2 | 256 | 2.00E-06 | 32 |
| llama2-13B_Bert_gold5 | 0.2 | 256 | 2.00E-06 | 2 |
| llama2-13B_Bert_gold50 | 0.2 | 256 | 2.00E-06 | 2 |
| llama2-13B_Bert_gold50_ws | 0.2 | 256 | 2.00E-06 | 32 |
| llama2-13B_Bert_gold5_ws | 0.2 | 256 | 2.00E-06 | 32 |
| llama2-13B_gold10_Bert_gold10 | 0.2 | 256 | 2.00E-06 | 2 |
| llama2-13B_gold10_Bert_gold10_ws | 0.2 | 256 | 2.00E-06 | 32 |
| llama2-13B_gold3_Bert_gold3 | 0.34 | 256 | 2.00E-06 | 1 |
| llama2-13B_gold3_Bert_gold3_ws | 0.2 | 256 | 2.00E-06 | 32 |
| llama2-13B_gold50_Bert_gold50 | 0.2 | 256 | 2.00E-06 | 2 |
| llama2-13B_gold50_Bert_gold50_ws | 0.2 | 256 | 2.00E-06 | 32 |
| llama2-13B_gold5_Bert_gold5 | 0.2 | 256 | 2.00E-06 | 2 |
| llama2-13B_gold5_Bert_gold5_ws | 0.2 | 256 | 2.00E-06 | 32 |
| 2018 (Track 2) ADE and Medication Extraction Challenge | | | | |
| Bert_gold10 | 0.2 | 256 | 2.00E-06 | 2 |
| Bert_gold100% | 0.2 | 256 | 2.00E-06 | 32 |
| Bert_gold3 | 0.34 | 256 | 2.00E-06 | 1 |
| Bert_gold5 | 0.2 | 256 | 2.00E-06 | 2 |
| Bert_gold50 | 0.2 | 256 | 2.00E-06 | 2 |
| llama2-13B_Bert_gold0 | 0.2 | 256 | 2.00E-06 | 32 |
| llama2-13B_Bert_gold10 | 0.2 | 256 | 2.00E-06 | 2 |
| llama2-13B_Bert_gold10_ws | 0.2 | 256 | 2.00E-06 | 32 |
| llama2-13B_Bert_gold5 | 0.2 | 256 | 2.00E-06 | 2 |
| llama2-13B_Bert_gold3_ws | 0.2 | 256 | 2.00E-06 | 32 |
| llama2-13B_Bert_gold5 | 0.2 | 256 | 2.00E-06 | 2 |
| llama2-13B_Bert_gold50 | 0.2 | 256 | 2.00E-06 | 2 |
| llama2-13B_Bert_gold50_ws | 0.2 | 256 | 2.00E-06 | 32 |
| llama2-13B_Bert_gold5_ws | 0.2 | 256 | 2.00E-06 | 32 |
| llama2-13B_gold10_Bert_gold10 | 0.2 | 256 | 2.00E-06 | 2 |
| llama2-13B_gold10_Bert_gold10_ws | 0.2 | 256 | 2.00E-06 | 32 |
| llama2-13B_gold3_Bert_gold3 | 0.34 | 256 | 2.00E-06 | 1 |
| llama2-13B_gold3_Bert_gold3_ws | 0.2 | 256 | 2.00E-06 | 32 |
| llama2-13B_gold50_Bert_gold50 | 0.2 | 256 | 2.00E-06 | 2 |
| llama2-13B_gold50_Bert_gold50_ws | 0.2 | 256 | 2.00E-06 | 32 |
| llama2-13B_gold5_Bert_gold5 | 0.2 | 256 | 2.00E-06 | 2 |
| llama2-13B_gold5_Bert_gold5_ws | 0.2 | 256 | 2.00E-06 | 32 |